\documentclass[conference]{IEEEtran}
\IEEEoverridecommandlockouts

\usepackage{cite}
\usepackage{amsmath,amssymb,amsfonts}
\usepackage{algorithmic}
\usepackage{graphicx}
\usepackage{textcomp}
\usepackage{xcolor}

\usepackage{multirow}
\usepackage{booktabs}
\usepackage{makecell}
\usepackage{xcolor}
\usepackage{pifont}
\usepackage{graphicx}
\usepackage{url}
\usepackage[table]{xcolor}

\def\BibTeX{{\rm B\kern-.05em{\sc i\kern-.025em b}\kern-.08em
    T\kern-.1667em\lower.7ex\hbox{E}\kern-.125emX}}
\begin{document}

\newcommand{\cmark}{\textcolor{green!70!black}{\checkmark}}
\newcommand{\xmark}{\textcolor{red}{\ding{55}}}


\title{MedUAG: Unified Understanding and Generation for Medical Multimodal Models}

\author{
Zijie Meng\textsuperscript{1,$\star$}\thanks{$^\star$Equal contribution.}, 
Yuncheng Zhang\textsuperscript{1,$\star$}, 
Hualiang Wang\textsuperscript{2,$\star$},
Yitian Tang\textsuperscript{1},
Xiaotang Gai\textsuperscript{1},\\
Chen Shen\textsuperscript{1},
Songtao Jiang\textsuperscript{1},
Shaosheng Cao\textsuperscript{3}, 
Jian Wu\textsuperscript{1}, 
Xian Wu\textsuperscript{4,\dag}, 
Zuozhu Liu\textsuperscript{1,\dag}\thanks{$^\dagger$Corresponding authors.} \\
\textsuperscript{1}\textit{Zhejiang University},
\textsuperscript{2}\textit{Hong Kong University of Science and Technology}, \\
\textsuperscript{3}\textit{Tsinghua University},
\textsuperscript{4}\textit{Tencent Jarvis Lab} \\
}


\maketitle

\begin{abstract}
    Recent Multimodal Large Language Models (MLLMs) are rapidly evolving into unified understanding and generation (UAG) frameworks. However, extending these unified paradigms to the medical domain is hindered by: the absence of comprehensive training and evaluation benchmarks, and the lack of broadly validated unified medical model. To address these gaps, we present a comprehensive foundation for medical UAG. First, we construct MedUAGCorpus, the largest unified medical understanding and generation dataset to date, comprising over 6 million instances across 14 imaging modalities. Second, we introduce MedUAGBench, a systematic benchmark that expands medical generation evaluation to 12 diverse tasks under standardized protocols. Finally, leveraging these resources, we develop MedUAG, an end-to-end trained unified medical model. Extensive experiments demonstrate that MedUAG achieves strong performance across a wide array of understanding and generation tasks, establishing a competitive baseline and paving the way for next-generation medical multimodal systems.
\end{abstract}


\section{Introduction}
\begin{figure}[t]
    \centering
    \includegraphics[width=\linewidth]{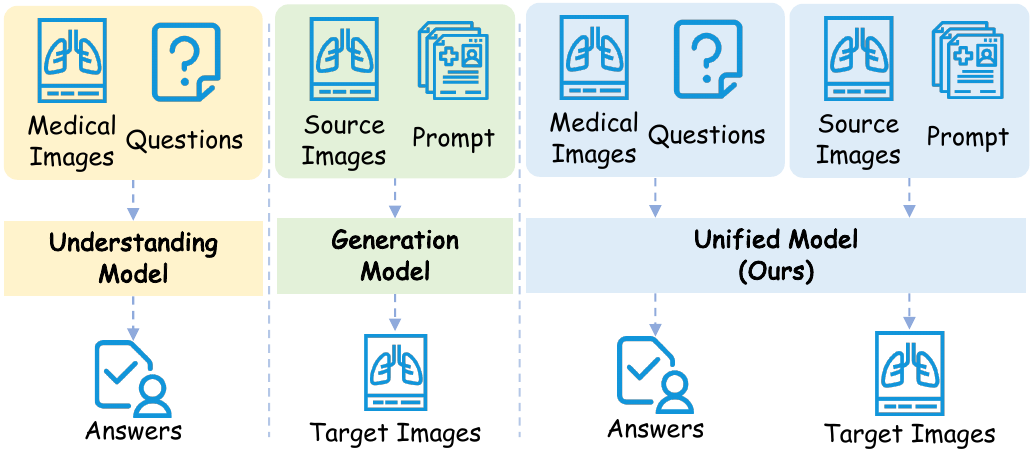}
    \caption{Comparison of different medical multimodal modeling paradigms. Our framework unifies understanding and generation within a single model.}
    \label{fig:toy_cmp}
\end{figure}

Recent Multimodal Large Language Models (MLLMs) have evolved from isolated understanding or generation systems into unified understanding and generation (UAG) frameworks across heterogeneous modalities~\cite{zhao2025unified}, as shown in Figure~\ref{fig:toy_cmp}. Driven by the scaling and integration of data across diverse UAG tasks, these unified training paradigms have demonstrated the potential to learn universal representations within a single system, substantially reducing the reliance on task-specific fine-tuning for downstream applications~\cite{ge2025seedxmultimodalmodelsunified,chen2025janus-pro,bagel,chen2025blip3,EMU3}.

However, the few pioneering works in medical UAG~\cite{lin2025healthgpt,ning2025unimedvl} are hindered by two significant limitations. \textbf{(1) The absence of comprehensive UAG training corpora and evaluation benchmarks.} While training and evaluating MLLMs necessitate large-scale data across diverse tasks, current studies often rely on a limited scope of medical imaging modalities and clinical applications. They resort to a narrow selection of common or readily accessible tasks, such as basic text-conditioned image synthesis, CT-MRI translation or MRI super resolution. Consequently, some challenging generation tasks of profound clinical significance remain overlooked. \textbf{(2) The lack of a broadly validated unified medical model.} Beyond data and benchmarks, current medical UAG studies have yet to establish a unified model that is trained on large-scale and diverse unified medical corpora and systematically validated across comprehensive understanding and generation benchmarks. Although recent approaches~\cite{lin2025healthgpt,ning2025unimedvl} have shown promising initial results, their empirical validation remains limited in task breadth, modality coverage, and evaluation consistency. As a result, it remains unclear whether a single end-to-end model can robustly handle diverse medical multimodal demands, ranging from semantic understanding and clinical reasoning to structurally faithful and clinically meaningful image generation. This limitation not only leaves the generality and robustness of unified medical modeling insufficiently demonstrated, but also makes it difficult to assess its potential for practical application.

\begin{table*}[t]
    \centering
    \caption{Comparison with existing unified medical understanding and generation works.}
    \begin{tabular}{l|cccccccc}
        \toprule
        {} & \multirow{2.5}{*}{{\makecell[c]{Modalities in\\Generation}}} & \multicolumn{2}{c}{{Understanding}} & \multicolumn{4}{c}{{Generation Tasks (\#)}} & \multirow{2.5}{*}{{Data Scale}} \\
        \cmidrule(lr){3-4} 
        \cmidrule(lr){5-8} 
        {} & {} & {VQA} & {MRG} & {Synthesis} & {Translation} & {Reconstruction} & {Prediction} & {} \\
        \midrule
        HealthGPT~\cite{lin2025healthgpt} & 11 & \cmark & \cmark & 1 & 2 & 2 & \xmark & 1.5M \\
        UniMedVL~\cite{ning2025unimedvl} & 10 & \cmark & \cmark & 2 & 1 & 1 & 1 & 5.6M \\
        MedUAG (ours)  & \textbf{14} & \cmark & \cmark & \textbf{3} & \textbf{3} & \textbf{4} & \textbf{2} & \textbf{6.4M} \\
        \bottomrule
    \end{tabular}
    \label{tab:bench_cmp}
\end{table*}

\begin{figure*}[t]
    \centering
    \includegraphics[width=\textwidth]{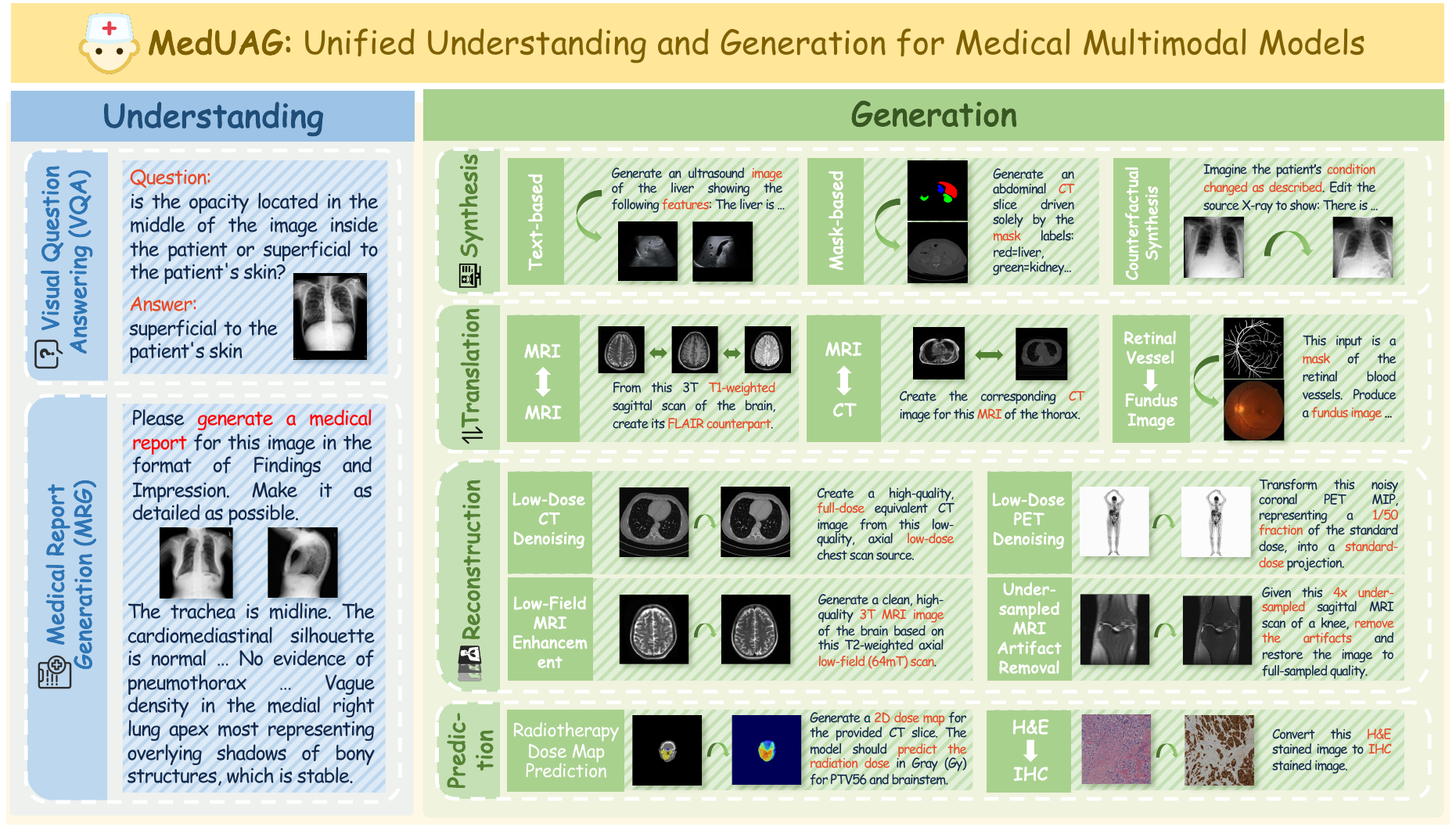}
    \caption{Overview of MedUAG. Our unified framework supports both medical understanding and generation. On the understanding side, it covers VQA and MRG. On the generation side, it supports four task categories: synthesis, translation, reconstruction, and prediction across diverse medical modalities. Representative demonstrations of each task type are provided.}
    \label{fig:overview}
\end{figure*}

To address these gaps, we introduce MedUAGBench, a comprehensive benchmark for unified medical generation. MedUAGBench extend the 5 tasks provided by UniMedVL~\cite{ning2025unimedvl} to 12 tasks in terms of the medical image generation, including various modalities under standardized prompts, metrics, and evaluation settings, enabling systematic and reproducible assessment of medical generative capability. In parallel, we construct MedUAGCorpus, the largest unified medical understanding and generation dataset to date, containing over 6M instances across 14 imaging modalities and diverse task types. Together, these resources provide the large-scale and diverse supervision foundation needed to support the comprehensive training and systematic validation of unified medical models, as summarized in Table~\ref{tab:bench_cmp} and Figure~\ref{fig:overview}.

Building upon these resources, we develop MedUAG, an end-to-end model for unified medical understanding and generation. Extensive experiments show that MedUAG achieves strong performance across a broad range of tasks, establishing a competitive baseline for future research. Overall, our work provides a unified foundation spanning benchmark, data corpus, and model, which takes a step toward more general and practically useful next-generation medical multi-modal systems.  Our main contributions are as follows: 
\begin{itemize}
    \item We introduce \textbf{MedUAGBench}, a comprehensive benchmark for unified medical generation, covering 12 tasks and various modalities under standardized evaluation protocols.
    \item We construct \textbf{MedUAGCorpus}, a large-scale unified understanding generation corpus with over 6M instances across 14 medical modalities and diverse task types.
    \item We develop \textbf{MedUAG}, an end-to-end unified medical model, and demonstrate strong performance across a wide range of understanding and generation tasks.
\end{itemize}

\begin{figure*}[t]
    \centering
    \includegraphics[width=\textwidth]{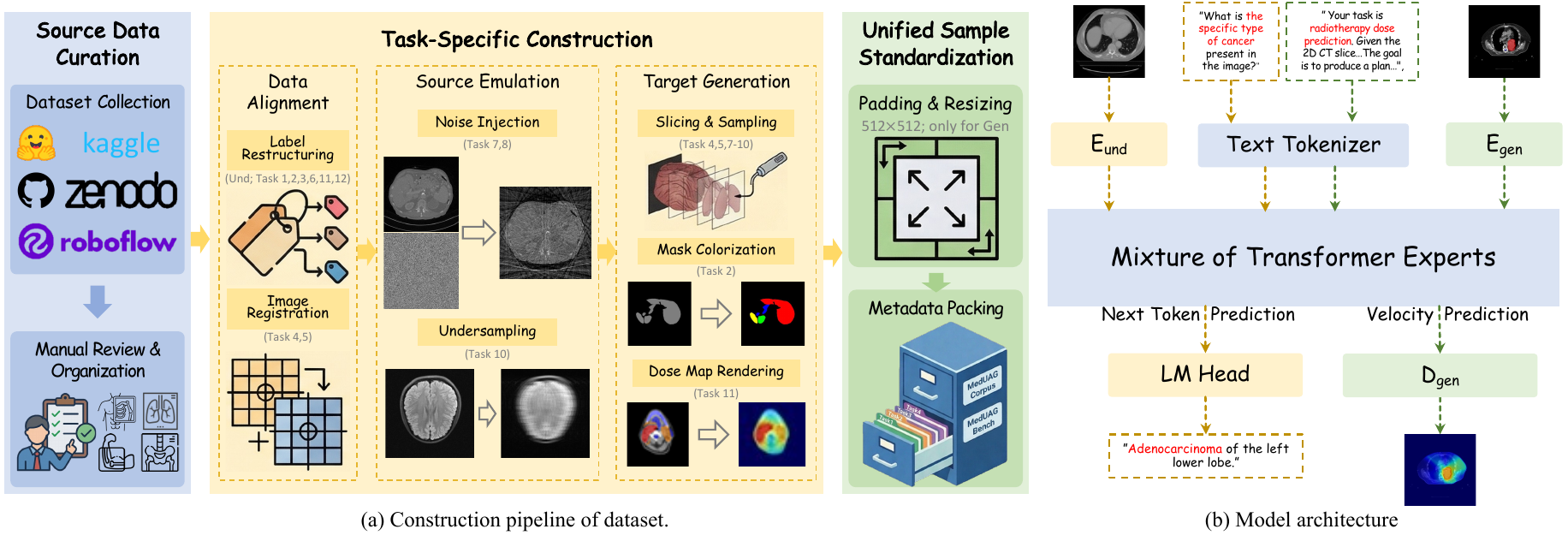}
    \caption{The construction process of datasets and model architecture of MedUAG. The dataset construction pipeline includes source data curation, task-specific construction, and unified sample standardization. The model adopts dual image encoders and mixture of transformer experts to support both medical image understanding and generation within a single framework.}
    \label{fig:pipeline}
\end{figure*}

\section{Methods}
\subsection{Definition of Tasks}
As illustrated in Figure~\ref{fig:overview}, we study unified medical multimodal modeling through two primary task domains: understanding and generation.

\subsubsection{Understanding Tasks}
Understanding tasks evaluate the model's ability to interpret medical images and extract clinically relevant semantics. We consider two primary tasks: (i) \textit{Visual Question Answering} (VQA), which requires answering natural language questions grounded in specific medical images and evaluates localized diagnostic reasoning and response precision; and (ii) \textit{Medical Report Generation} (MRG), which requires generating a complete clinical report from medical images. Unlike VQA, MRG requires the model to identify clinically significant findings and summarize them in a structured and coherent form.

\subsubsection{Generation Tasks}
Generation tasks evaluate the model's ability to produce medically meaningful images under diverse conditions. We categorize them into four paradigms. (i) \textit{Synthesis} creates new medical images or edits existing ones under semantic guidance, including text-based, mask-based, and counterfactual settings. This requires the model to capture anatomical structure and pathological semantics, enabling plausible image generation or clinically specified image modification. Such capability is useful for data augmentation, medical education, and visualization of potential disease progression. (ii) \textit{Translation} transforms images across domains while preserving the underlying anatomical content. It includes intra-modality translation, inter-modality translation, and structure-to-image generation, evaluating whether the model can preserve shared structures while adapting domain-specific appearance. Clinically, it can synthesize complementary views or modalities from existing scans, enriching diagnostic information without additional acquisition. (iii) \textit{Reconstruction} recovers high-quality images from degraded inputs. In our dataset, this category includes low-dose CT denoising, low-dose PET denoising, low-field MRI enhancement, and undersampled MRI restoration. Successful reconstruction requires preserving diagnostically relevant details while reducing acquisition-induced distortions, supporting more efficient and lower-risk imaging protocols. (iv) \textit{Prediction} generates clinically informative outputs beyond direct restoration or cross-domain translation. This category includes radiotherapy dose distribution prediction from CT images with annotated target volumes and organs at risk, as well as H\&E-to-IHC pathological stain conversion. Compared with other generation tasks, prediction emphasizes outputs with more direct clinical relevance and application potential.

\begin{figure*}[t]
    \centering
    \includegraphics[width=\textwidth]{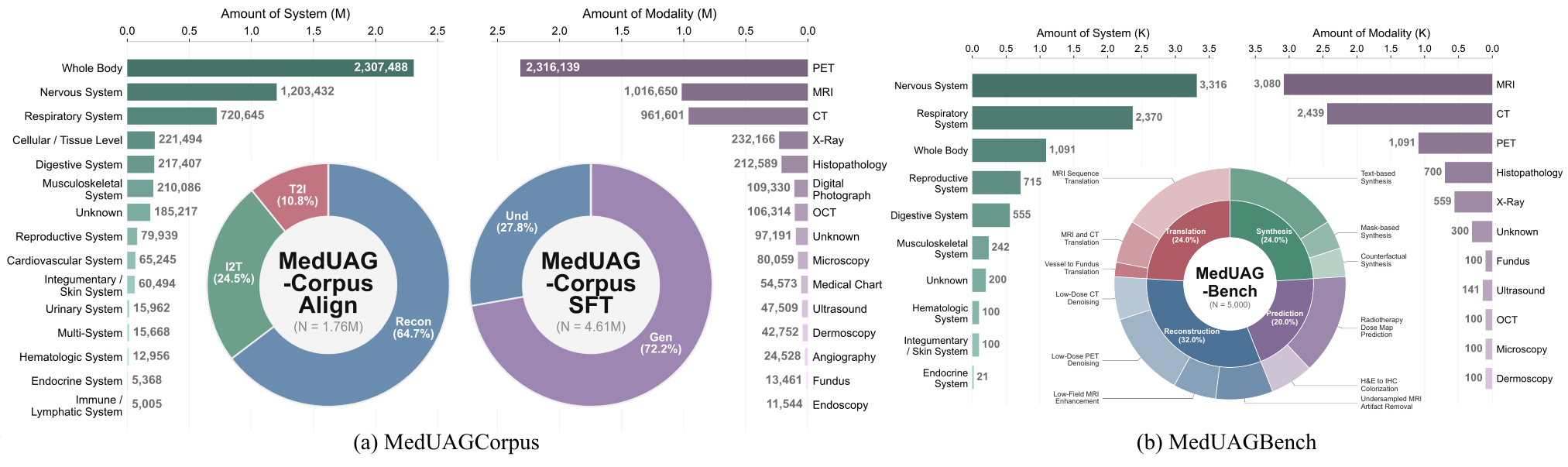}
    \caption{Statistical overview of dataset. (a) MedUAGCorpus: distribution of samples across anatomical systems and imaging modalities, together with the composition of the two training stages, including MedUAG-Corpus Align and MedUAG-Corpus SFT. ``T2I'', ``I2T'', ``Recon'', ``Und'' and ``Gen'' indicate text-to-image, image-to-text, reconstruction, understanding and generation, respectively. (b) MedUAGBench: distribution of benchmark samples across anatomical systems and imaging modalities, along with the hierarchical task composition covering reconstruction, translation, synthesis, and prediction.}
    \label{fig:data_statistic}
\end{figure*}

\subsection{Construction of Datasets}
As illustrated in Figure~\ref{fig:pipeline}, we construct a systematic and extensible pipeline to build the unified corpus and benchmark.

\subsubsection{Source Data Curation}
We first collect public datasets according to the task taxonomy, and then manually review and normalize them into a unified data pool for task-aware processing. For understanding tasks, we mainly use medical multimodal data from Hulu-Med~\cite{jiang2025hulumedtransparentgeneralistmodel}. For generation tasks, we collect public datasets covering synthesis, translation, reconstruction, and prediction~\cite{chambon2024chexpertplusaugmentinglarge,Yang_2023,xie2025medtrinity25mlargescalemultimodaldataset,R_ckert_2024,9497733,ma2025interpretablecounterfactualgenerationmultimodal,ulfenc2025,ixi_dataset,Thummerer_2025,jin2022fives,1282003,Moen2021,Armato2015LIDC,XueSon_UDPET_MICCAI2025,zbontar2019fastmriopendatasetbenchmarks,liu2022bcibreastcancerimmunohistochemical,li2023adaptivesupervisedpatchnceloss,Babier_2021,gao2025automatingrtplanningscale,gdp_hmm2025}. After collection, all datasets are manually inspected to remove unusable samples and harmonize file structures and annotation formats, resulting in normalized raw data for subsequent construction.

\subsubsection{Task-Specific Construction} 
Following curation, task-specific construction further exploits the information available in each dataset and derives samples that match different downstream scenarios. This stage consists of three components: (i) Data alignment includes label restructuring and image registration, which adapt existing annotations to task-specific requirements and, when necessary, align images of the same anatomical structures across different modalities. (ii) Source emulation simulates degraded acquisition conditions, such as noise injection and undersampling. (iii) Target generation performs task-dependent operations, including slicing and sampling volumetric data into 2D images, mask colorization for mask-based synthesis, and dose map rendering for radiotherapy dose prediction. Through these procedures, heterogeneous raw data are transformed into unified input-output pairs for different tasks.

\subsubsection{Unified Sample Standardization} 
Finally, all processed samples are standardized into a unified format and split into the training corpus and benchmark at the patient level. Specifically, all images, slices, and derived samples from the same patient are assigned exclusively to either the training corpus or the benchmark, preventing data leakage between training and evaluation. Furthermore, to satisfy the fixed input size required by some generative models, images in generation tasks are resized and padded to $512 \times 512$ while preserving the original anatomical aspect ratio as much as possible. Each sample is then packaged with its image path, metadata, unique identifier, and task-specific text prompt. This step ensures format consistency across samples and supports unified multi-task training and evaluation.

\subsubsection{Dataset Statistics}
As shown in Figure~\ref{fig:data_statistic}, our dataset exhibits broad diversity in both the corpus and benchmark. MedUAGCorpus contains over 6M training instances, including a 1.76M domain-alignment set and a 4.61M instruction-tuning set. It comprises more than 5.3M images and spans 14 anatomical systems and 14 imaging modalities. In contrast, MedUAGBench is more compact and evaluation-oriented. Since it focuses primarily on medical generation, its system and modality coverage is smaller, yet it still spans 10 anatomical systems and 11 imaging modalities. Moreover, the benchmark remains task-diverse, covering four core generation categories, thereby enabling comprehensive and fine-grained evaluation of medical image generation.


\subsection{Implementation of MedUAG}
\paragraph{Model Architecure}
As shown in Figure~\ref{fig:pipeline}, MedUAG follows the design of Bagel~\cite{bagel} to balance the different granularity requirements of understanding and generation within a unified architecture. Specifically, it uses a ViT encoder~\cite{vit} to extract high-level semantic visual tokens for understanding, and a VAE encoder-decoder~\cite{vae} to model low-level latent representations for image generation. Built on a decoder-only transformer backbone~\cite{bai2025qwen2.5vl}, the model instantiates two task-specific branches for understanding and generation, decoupling their optimization while preserving architectural consistency.

For understanding, the model takes text tokens and ViT visual tokens as input and performs autoregressive next-token prediction, following standard MLLMs~\cite{llava,bai2025qwen2.5vl}. The objective is the cross-entropy loss:
\begin{equation}
\mathcal{L}_{\mathrm{und}}
=
-\mathbb{E}_{(\mathbf{x},\mathbf{v},\mathbf{y}) \sim \mathcal{D}}
\left[
\sum_{t=1}^{|\mathbf{y}|}
\log p_{\theta}\!\left(y_t \mid \mathbf{x}, \mathbf{v}, y_{<t}\right)
\right],
\end{equation}
where $\mathbf{x}$ denotes the input text tokens, $\mathbf{v}$ the ViT visual tokens, and $\mathbf{y}$ the target output sequence.

For generation, the model operates in the VAE latent space and predicts the target velocity under the flow-matching objective~\cite{liu2022flow,lipman2022flow}:
\begin{equation}
\mathcal{L}_{\mathrm{gen}}
=
\mathbb{E}_{(\mathbf{z}_t,\mathbf{u}_t,\mathbf{c}) \sim \mathcal{D}}
\left[
\left\|
f_{\theta}(\mathbf{z}_t, t, \mathbf{c}) - \mathbf{u}_t
\right\|_2^2
\right],
\end{equation}
where $\mathbf{z}_t$ is the noised latent feature at timestep $t$, $\mathbf{c}$ is the conditioning information, and $\mathbf{u}_t$ is the target velocity.

The final text and image outputs are decoded by the LM head and VAE decoder, respectively. The overall objective is:
\begin{equation}
\mathcal{L}
=
\lambda_{\mathrm{und}} \mathcal{L}_{\mathrm{und}}
+
\lambda_{\mathrm{gen}} \mathcal{L}_{\mathrm{gen}},
\end{equation}
where $\lambda_{\mathrm{und}}$ and $\lambda_{\mathrm{gen}}$ balance these two losses.

\paragraph{Domain Alignment}
MedUAG is trained in two stages. The first stage adapts the pretrained backbone to the medical domain before unified instruction tuning. We construct the alignment corpus from three representative tasks: medical image reconstruction, text-to-image generation, and image captioning. These tasks provide complementary supervision: reconstruction offers dense pixel-level guidance for the generation branch, text-to-image generation strengthens medical text-guided synthesis, and captioning improves image-language semantic alignment for the understanding branch. This stage establishes domain-aware representations and provides a stable initialization for subsequent instruction tuning.

\paragraph{Instruction Tuning}
The second stage trains MedUAG with unified multimodal instructions covering both generation and understanding. The generation data include synthesis, translation, reconstruction, and prediction, while the understanding data include VQA and MRG. Compared with domain alignment, this stage emphasizes task diversity, compositional conditioning, and instruction responsiveness. Through joint instruction tuning, the understanding branch learns language-guided reasoning and reporting, while the generation branch learns diverse conditional image generation objectives, yielding a general-purpose medical multimodal model.

\begin{table*}[t]
    \centering
    \caption{Average results across the four generation categories. For synthesis, we report task-level macro averages of FID, gFID, and BioCS. For translation, reconstruction, and prediction, we report task-level macro averages of LPIPS, MSE, PSNR, and SSIM.}
    \resizebox{\linewidth}{!}{%
    \begin{tabular}{lc|ccc|cccc|cccc|cccc}
        \toprule
        {} & \multirow{2.5}{*}{Medical}
        & \multicolumn{3}{c|}{{Synthesis Avg.}}
        & \multicolumn{4}{c|}{{Translation Avg.}}
        & \multicolumn{4}{c|}{{Reconstruction Avg.}}
        & \multicolumn{4}{c}{{Prediction Avg.}} \\
        \cmidrule(lr){3-5}
        \cmidrule(lr){6-9}
        \cmidrule(lr){10-13}
        \cmidrule(lr){14-17}
        {} & {}
        & {FID$\downarrow$} & {gFID$\downarrow$} & {BioCS$\uparrow$}
        & {LPIPS$\downarrow$} & {MSE$\downarrow$} & {PSNR$\uparrow$} & {SSIM$\uparrow$}
        & {LPIPS$\downarrow$} & {MSE$\downarrow$} & {PSNR$\uparrow$} & {SSIM$\uparrow$}
        & {LPIPS$\downarrow$} & {MSE$\downarrow$} & {PSNR$\uparrow$} & {SSIM$\uparrow$} \\
        \midrule
        BLIP3-o~\cite{chen2025blip3} & \xmark
        & 320.35 & 316.74 & 0.349
        & 0.625 & 0.064 & 12.536 & 0.348
        & 0.513 & 0.056 & 13.177 & 0.373
        & 0.713 & 0.091 & 10.596 & 0.268 \\
        UniWorld-V1~\cite{lin2025uniworldv1highresolutionsemanticencoders} & \xmark
        & 334.51 & 332.54 & 0.321
        & 0.609 & \textbf{0.053} & 13.141 & 0.377
        & 0.602 & 0.069 & 11.931 & 0.287
        & 0.674 & 0.084 & 11.140 & 0.150 \\
        Bagel~\cite{bagel} & \xmark
        & 261.39 & 248.72 & 0.361
        & 0.613 & 0.186 & 9.147 & 0.307
        & 0.523 & 0.123 & 12.301 & 0.329
        & 0.602 & 0.165 & 8.496 & 0.249 \\
        \midrule
        HealthGPT~\cite{lin2025healthgpt} & \cmark
        & 211.63 & 202.48 & 0.381
        & 0.466 & 0.057 & 13.659 & 0.416
        & 0.370 & 0.024 & 17.000 & 0.533
        & 0.679 & 0.076 & 12.003 & 0.262 \\
        UniMedVL~\cite{ning2025unimedvl} & \cmark
        & 192.21 & 179.32 & \textbf{0.403}
        & 0.407 & 0.054 & 13.616 & 0.486
        & 0.365 & 0.063 & 13.439 & 0.455
        & 0.531 & 0.129 & 11.237 & 0.354 \\
        \rowcolor{blue!5} 
        MedUAG (ours) & \cmark
        & \textbf{181.72} & \textbf{167.72} & 0.396
        & \textbf{0.364} & 0.063 & \textbf{16.992} & \textbf{0.575}
        & \textbf{0.113} & \textbf{0.007} & \textbf{25.357} & \textbf{0.809}
        & \textbf{0.320} & \textbf{0.039} & \textbf{17.897} & \textbf{0.529} \\
        \bottomrule
    \end{tabular}%
    }
    \label{tab:overall_gen_avg}
\end{table*}

\section{Experiments}
\subsection{Implementation Details}
We train MedUAG in two stages initializing from the pretrained BAGEL-7B-MoT. In the alignment stage, the model is trained for 10k steps with a learning rate of $5\times10^{-5}$ on 1.76M samples, including 1.14M reconstruction samples (64.6\%), 191K text-to-image samples (10.8\%), and 432K image-to-text samples (24.5\%). In the subsequent SFT stage, training is resumed from the aligned checkpoint and continued for 20k steps with a learning rate of $2\times10^{-5}$ on 4.61M instruction-tuning samples, consisting of 3.33M generation samples (72.3\%) and 1.28M single-image understanding samples (27.7\%). Training is conducted on 32 NVIDIA H800 80GB GPUs. In both stages, we set the maximum number of tokens per sample to 16,384, and adopt a constant learning-rate schedule with 2,000 warmup steps, the AdamW optimizer ($\beta_1=0.9$, $\beta_2=0.95$, $\epsilon=10^{-15}$), EMA decay of 0.9999, and frozen VAE weights. The loss weights $\lambda_{\mathrm{und}}$ and $\lambda_{\mathrm{gen}}$ are both set to 1.0.

\begin{table}[t]
    \centering
    \caption{Comparison among various models on understanding benchmarks, where OM.VQA indicates OmniMedVQA.}
    \resizebox{\linewidth}{!}{%
    \begin{tabular}{lc|cccc|c}
        \toprule
        Model & Unified
        & VQA-RAD
        & SLAKE
        & PathVQA
        & OM.VQA
        & Avg. \\
        \midrule
        \multicolumn{7}{c}{\textit{Proprietary Baselines}} \\
        \midrule
        GPT-4.1 & \xmark & 65.0 & 72.2 & 55.5 & 75.5 & 67.1 \\
        Claude Sonnet 4 & \xmark & 67.6 & 70.6 & 54.2 & 65.5 & 64.5 \\
        Gemini-2.5-Flash & \xmark & 68.5 & 75.8 & 55.4 & 71.0 & 67.7 \\
        \midrule
        \multicolumn{7}{c}{\textit{General-purpose Baslines}} \\
        \midrule
        Qwen2.5-VL-32B & \xmark & 71.8 & 71.2 & 41.9 & 68.2 & 63.3 \\
        InternVL3-38B & \xmark & 65.4 & 72.7 & 51.0 & 79.8 & 67.2 \\
        Llama3.2-11B & \xmark & 58.8 & 65.8 & 32.9 & 43.8 & 50.3 \\
        Qwen2.5VL-7B & \xmark & 63.2 & 66.8 & 44.1 & 63.6 & 59.4 \\
        InternVL3-8B & \xmark & 65.4 & 72.8 & 48.6 & 79.1 & 66.5 \\
        Janus-Pro-7B & \cmark & 49.7 & 55.2 & 35.4 & 59.6 & 50.0 \\
        Bagel & \cmark & 60.1 & 58.9 & 39.1 & 71.1 & 57.3 \\
        \midrule
        \multicolumn{7}{c}{\textit{Medical Baselines}} \\
        \midrule
        LLaVA-Med-7B & \xmark & 46.6 & 51.9 & 35.2 & 34.8 & 42.1 \\
        RadFM & \xmark & 50.6 & 34.6 & 14.3 & 23.5 & 30.8 \\
        HuatuoGPT-V-7B & \xmark & \underline{67.6} & 68.1 & 44.8 & 74.3 & 63.7 \\
        HealthGPT-M3 & \cmark & 55.9 & 56.4 & 39.7 & 68.5 & 55.1 \\
        HealthGPT-L14 & \cmark & 58.3 & 64.5 & 44.4 & 74.4 & 60.4 \\
        UniMedVL & \cmark & 61.9 & \underline{75.4} & \underline{53.5} & \textbf{85.8} & \underline{69.2} \\
        \rowcolor{blue!5} 
        MedUAG (ours) & \cmark & \textbf{75.6} & \textbf{78.0} & \textbf{54.5} & \underline{77.0} & \textbf{71.3} \\
        \bottomrule
    \end{tabular}%
    }
    \label{tab:overall_und_avg}
\end{table}

\subsection{Benchmarks, Baselines and Metrics}
\paragraph{Benchmarks}
We evaluate MedUAG on both medical generation and understanding tasks to comprehensively assess its unified capability. For generation, we use MedUAGBench, a unified medical generation benchmark comprising 5,000 test instances over 12 tasks, which cover diverse medical imaging scenarios and require the model to handle heterogeneous generation objectives. For understanding, we evaluate on four widely used medical benchmarks, including VQA-RAD~\cite{vqa-rad}, SLAKE~\cite{liu2021slake}, PathVQA~\cite{he2020pathvqa}, and OmniMedVQA~\cite{hu2024omnimedvqa}. Together, these benchmarks provide a broad assessment of medical visual understanding ability, covering complementary reasoning demands across radiology, pathology, and general clinical question answering.

\paragraph{Baselines}
We compare MedUAG with a diverse set of baselines covering both unified multimodal models and specialized medical MLLMs. For generation, we include representative unified multimodal models from the general domain, including BLIP3-o~\cite{chen2025blip3}, UniWorld-V1~\cite{lin2025uniworldv1highresolutionsemanticencoders}, and Bagel~\cite{bagel}, as well as medical-domain unified models, including HealthGPT~\cite{lin2025healthgpt} and UniMedVL~\cite{ning2025unimedvl}. For understanding,  we consider three groups of baselines: proprietary systems (GPT-4.1~\cite{openai_gpt41_2025}, Claude Sonnet 4~\cite{anthropic_claude4_2025}, and Gemini-2.5-Flash~\cite{comanici2025gemini2.5}), general-purpose open-source models (\textit{e.g.}, Qwen2.5-VL~\cite{bai2025qwen2.5vl}, InternVL3~\cite{zhu2025internvl3}, Llama3.2~\cite{grattafiori2024llama}, Janus-Pro~\cite{chen2025janus-pro}, and Bagel~\cite{bagel}), and specialized medical models (\textit{e.g.}, LLaVA-Med~\cite{llava-med}, RadFM~\cite{radfm}, HuatuoGPT-V~\cite{huatuogptvl}, HealthGPT~\cite{lin2025healthgpt}, and UniMedVL~\cite{ning2025unimedvl}).

\begin{figure}[t]
    \centering
    \includegraphics[width=\linewidth]{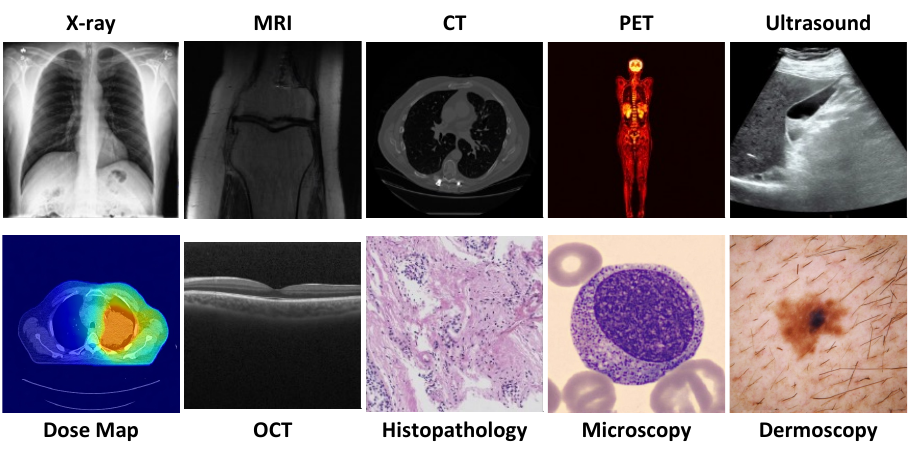}
    \caption{Visualization of the MedUAG generation capability.}
    \label{fig:visualization}
\end{figure}

\paragraph{Metrics}
For MedUAGBench, we follow the standard evaluation protocol for each generation category. Specifically, for synthesis, we report FID~\cite{heusel2017gans}, generation FID (gFID), and BiomedCLIP Score (BioCS)~\cite{zhang2023biomedclip} to evaluate image realism, generative fidelity, and medical semantic consistency, respectively. For translation, reconstruction, and prediction, we report LPIPS~\cite{zhang2018unreasonable}, MSE, PSNR, and SSIM~\cite{wang2004image} to measure perceptual similarity, pixel-wise error, reconstruction quality, and structural consistency. For medical understanding benchmarks, we use accuracy as the evaluation metric, where open-ended questions are assessed by Qwen3-VL-30B-A3B-Instruct~\cite{bai2025qwen3vl}.

\subsection{Main Results}
\subsubsection{Comparison on MedUAGBench}
Table~\ref{tab:overall_gen_avg} reports the quantitative results on MedUAGBench. MedUAG consistently outperforms both general-domain unified models and prior medical unified models across most generation settings, demonstrating the benefit of large-scale medical unified training. Its advantage is especially clear on reconstruction and prediction, where preserving anatomical structure and producing clinically grounded outputs are essential. These results suggest that MedUAG does not merely improve image realism, but also learns generation capabilities aligned with medical structure and task semantics. The qualitative examples in Figure~\ref{fig:visualization} further show that MedUAG produces realistic and structurally faithful outputs across diverse imaging modalities.

\subsubsection{Comparison on Understanding Benchmarks}
Table~\ref{tab:overall_und_avg} reports results on medical understanding benchmarks. MedUAG achieves the best average accuracy of 71.3 against 16 baselines, leading on VQA-RAD and SLAKE with scores of 75.6 and 78.0, respectively, while remaining competitive on PathVQA and OmniMedVQA. Notably, although MedUAG is jointly trained for both generation and understanding, with understanding data comprising less than 30\% of the corpus in both stages, it still demonstrates strong visual reasoning across radiology, pathology, and general medical QA. Its gap to UniMedVL on OmniMedVQA suggests that data composition, the understanding-generation mixing ratio, and task balancing remain important directions for future unified medical multimodal models.

\subsection{Ablation Studies}
\begin{table}[t]
    \centering
    \caption{Ablation study of domain alignment and initialization.}
    \setlength{\tabcolsep}{5.5pt}
    \renewcommand{\arraystretch}{1.15}
    \resizebox{.9\linewidth}{!}{%
    \begin{tabular}{cc|cccc}
        \toprule
        \multirow{2}{*}{Align.} & \multirow{2}{*}{Init.} & Synthesis & Translation & Reconstruction & Prediction \\
        & & (gFID$\downarrow$) & (LPIPS$\downarrow$) & (PSNR$\uparrow$) & (MSE$\downarrow$) \\
        \midrule
        \xmark & Scratch & 332.40 & 0.478 & 22.449 & 0.053 \\
        \xmark & Bagel   & \textbf{128.28} & 0.396 & 24.476 & 0.040 \\
        \midrule
        \cmark & Scratch & 309.50 & 0.475 & 23.078 & 0.057 \\
        \cmark & Bagel   & 167.72 & \textbf{0.364} & \textbf{25.357} & \textbf{0.039} \\
        \bottomrule
    \end{tabular}%
    }
    \label{tab:ablation_alignment_init}
\end{table}

\subsubsection{Effect of Domain Alignment}
\begin{table}[t]
    \centering
    \caption{PSNR gains from domain alignment on reconstruction tasks.}
    \resizebox{\linewidth}{!}{%
    \begin{tabular}{c|cccc|c}
        \toprule
        Init.
        & \makecell[c]{Low-Dose\\CT}
        & \makecell[c]{Low-Dose\\PET}
        & \makecell[c]{Low-Field\\MRI}
        & \makecell[c]{Undersampled\\MRI}
        & {Avg.} \\ 
        \midrule
        From Scratch & +1.13 & +1.00 & +0.33 & +0.05 & +0.63 \\
        From Bagel   & +1.14 & +0.05 & +1.89 & +0.44 & +0.88 \\
        \bottomrule
    \end{tabular}%
    }
    \label{tab:alignment_recon_gain}
\end{table}

To assess domain alignment, we compare models with and without this stage under both from-scratch and Bagel initializations, as shown in Table~\ref{tab:ablation_alignment_init}. Domain alignment mainly benefits structure-preserving tasks: with Bagel initialization, it lowers translation LPIPS by $0.032$ and improves reconstruction PSNR by $0.881$, with consistent gains across reconstruction benchmarks in Table~\ref{tab:alignment_recon_gain}. However, synthesis performance decreases after alignment, revealing a trade-off between anatomical fidelity and generative diversity. This suggests that alignment-induced structural priors should be retained, while synthesis-oriented objectives should be progressively strengthened in later training to better balance structural fidelity and generative realism.

\subsubsection{Effect of Base Model}
Table~\ref{tab:ablation_alignment_init} further illustrates the efficacy of different initialization strategies. Compared to training from scratch, Bagel initialization consistently yields substantial gains across nearly all tasks, most notably in synthesis and reconstruction. Without domain alignment, Bagel drastically reduces gFID from $332.40$ to $128.28$ and boosts reconstruction PSNR by $2.027$. This superiority persists even after domain alignment is introduced, where Bagel still provides the optimal foundation for translation, reconstruction, and prediction. These results underscore that general-domain unified pretraining establishes a robust representational prior, which significantly lowers the barrier for downstream medical adaptation and ensures a better optimization starting point.

\subsubsection{Effect of Data Scale}
\begin{figure}[t]
    \centering
    \includegraphics[width=\linewidth]{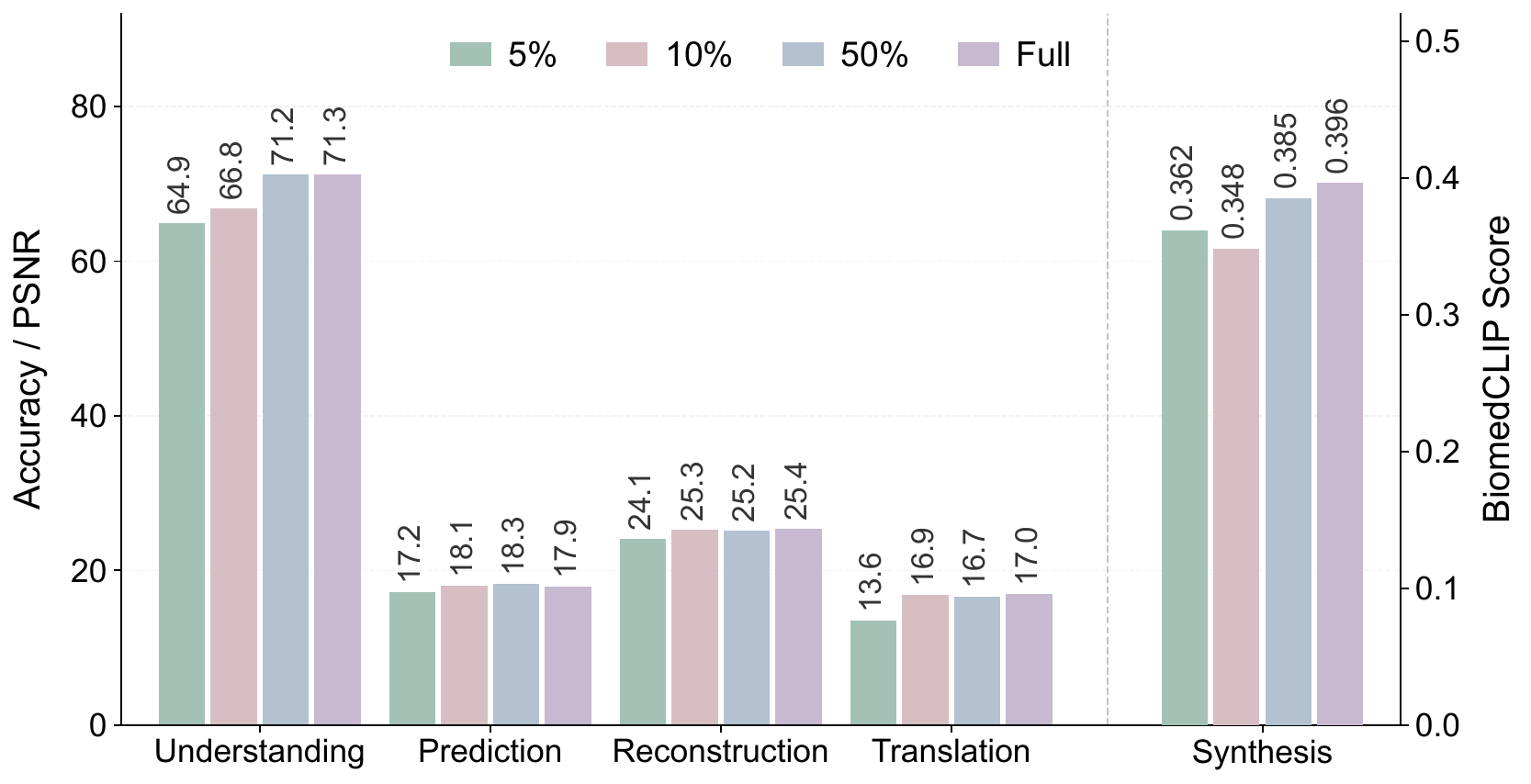}
    \caption{Comparison of different SFT data ratios.}
    \label{fig:ablation_data_scale}
\end{figure}

We further analyze the effect of SFT data scale as shown in Figure~\ref{fig:ablation_data_scale}. Increasing instruction-tuning data improves performance across all tasks, but the gains gradually saturate at larger scales. Understanding and synthesis show more sustained improvements than reconstruction and prediction, likely because they occupy smaller portions of the SFT set and therefore benefit more from increased diversity and coverage as the total data grows. These results suggest that, beyond sufficient scale, further gains depend more on data diversity, task balance, and quality control than on simply adding more samples.

\section{Application Exploration}
\begin{table}[t]
    \centering
    \caption{Effect of synthetic data on medical report generation.}
    \setlength{\tabcolsep}{6pt}
    \renewcommand{\arraystretch}{1.12}
    \resizebox{.95\linewidth}{!}{%
    \begin{tabular}{ccc|cccc}
        \toprule
        \#Real & \#Synthetic & \#Total & METEOR & BERTScore & LLM Judge & Average \\
        \midrule
        1K & 0    & 1K & 0.098 & 0.394 & 0.037 & 0.176 \\
        1K & 1K & 2K & 0.191 & 0.495 & 0.100 & 0.262 \\
        2K & 0    & 2K & 0.196 & 0.491 & 0.125 & 0.271 \\
        \bottomrule
    \end{tabular}%
    }
    \label{tab:application}
\end{table}

Beyond unified understanding and generation, MedUAG can also synthesize multimodal data for downstream training in low-resource settings, where paired clinical data are often scarce, costly, or privacy-restricted~\cite{wang2025self}. To evaluate this potential, we conduct an exploratory experiment on MRG task from CheXpert~\cite{irvin2019chexpert} using Qwen3-VL-4B as the backbone. We compare three training settings: 1K real image-report pairs, 1K real pairs augmented with 1K synthetic images generated by MedUAG from the corresponding reports, and a 2K-real upper bound. As shown in Table~\ref{tab:application}, synthetic augmentation improves the 1K-real baseline by 0.086 on average and narrows the gap to the 2K-real upper bound to only 0.009. These results suggest that MedUAG-generated images can provide effective supervision when real paired data are limited, highlighting the potential of unified medical multimodal models for scalable and cost-effective data augmentation.

\section{Related Work}
Recent medical vision-language models adapt general-purpose MLLMs~\cite{bai2025qwen2.5vl,bai2025qwen3vl,grattafiori2024llama} to clinical scenarios~\cite{llava-med,chen2024towards,radfm,jiang2025hulu,Zhang_2024}. These models have shown strong capabilities in medical VQA, report generation, and clinical reasoning by leveraging medical image-text alignment, instruction tuning, and large-scale domain-specific data. However, most Med-MLLMs are still centered on visual comprehension and text generation. They cannot directly synthesize, translate, reconstruct, or manipulate medical images at the pixel level, limiting their applicability to generative clinical tasks.

UAG has recently become an active direction in general-domain multimodal learning~\cite{chen2025blip3,EMU3,chen2025janus-pro,lin2025uniworldv1highresolutionsemanticencoders,bagel}. Existing methods explore different design choices, including coupling LLMs with generative modules, discretizing images into visual tokens, decoupling visual encoders, and adopting expert-based architectures to support both perception and generation. Despite their strong open-domain generalization, these models are primarily trained on natural images. As a result, they often struggle with clinical semantics, anatomical fidelity, and fine-grained pathology in medical imaging.

Recent works have begun to explore unified multimodal models for medicine. HealthGPT~\cite{lin2025healthgpt} unifies comprehension and generation with heterogeneous knowledge adaptation, and UniMed-VL~\cite{ning2025unimedvl} models further mark important progress toward unified medical AI. Nevertheless, existing efforts remain constrained by limited data scale, narrow generation task coverage, and insufficiently standardized evaluation. In contrast, our work provides a more comprehensive foundation by constructing a large-scale corpus, a systematic generation benchmark, and an end-to-end unified medical model.

\section{Conclusion}
We present MedUAG, a unified medical multimodal model, together with MedUAGBench and MedUAGCorpus for systematic training and evaluation of medical understanding and generation. MedUAG achieves strong performance across both task families, with clear gains in reconstruction and prediction, demonstrating the promise of unified modeling for general-purpose medical AI. We also identify key challenges, including limitations in some understanding settings and a trade-off between structural fidelity and synthesis diversity. Future work will further improve data composition, task balancing, and training strategies toward more robust and clinically useful medical multimodal systems.

\bibliographystyle{IEEEtran}
\bibliography{reference}

\end{document}